\documentclass[10pt,twocolumn,letterpaper]{article}

\usepackage[numbers]{natbib}
\usepackage{cvpr}
\usepackage{times}
\usepackage{epsfig}
\usepackage{graphicx}
\usepackage{amsmath}
\usepackage{amssymb}
\usepackage{multirow}
\usepackage{bm}
\usepackage{booktabs}
\usepackage{subfigure}
\usepackage[colorlinks,linkcolor=blue]{hyperref}

\cvprfinalcopy 

\def\cvprPaperID{****} 

\ifcvprfinal\fi
\begin{document}

\title{HYouTube: Video Harmonization Dataset}

\author{$\textnormal{Xinyuan Lu}$, $\textnormal{Shengyuan Huang}$, $\textnormal{Li Niu}$, $\textnormal{Wenyan Cong}$, $\textnormal{Liqing Zhang}$\\
Shanghai Jiao Tong University \\
}

\maketitle
\thispagestyle{empty}

\begin{abstract}
Video composition aims to generate a composite video by combining the foreground of one video with the background of another video, but the inserted foreground may be incompatible with the background in terms of color and illumination. 
Video harmonization aims to adjust the foreground of a composite video to make it compatible with the background. So far, video harmonization has only received limited attention and there is no public dataset for video harmonization. In this work, we construct a new video harmonization dataset HYouTube by adjusting the foreground of real videos to create synthetic composite videos. Considering the domain gap between real composite videos and synthetic composite videos, we additionally create 100 real composite videos via copy-and-paste. 
Datasets are available at \href{https://github.com/bcmi/Video-Harmonization-Dataset-HYouTube}{https://github.com/bcmi/Video-Harmonization-Dataset-HYouTube}.

\end{abstract}


\section{Introduction}

Image or video composition is a common operation to create visual content. Given two different videos, video composition aims to generate a composite video by combining the foreground of one video with the background of another video. However, composite videos are usually not realistic enough due to the appearance (\emph{e.g.}, illumination, color) incompatibility between foreground and background, which is caused by distinctive capture conditions (\emph{e.g.}, season, weather, time of the day) of foreground and background \cite{2020DoveNet,2021Bargainnet}. To address this issue, video harmonization \cite{2019Temporally} has been proposed to adjust the foreground appearance to make it compatible with the background, resulting in a more realistic composite video.

\begin{figure}[t]
    \centering
    \includegraphics[width=0.45\textwidth]{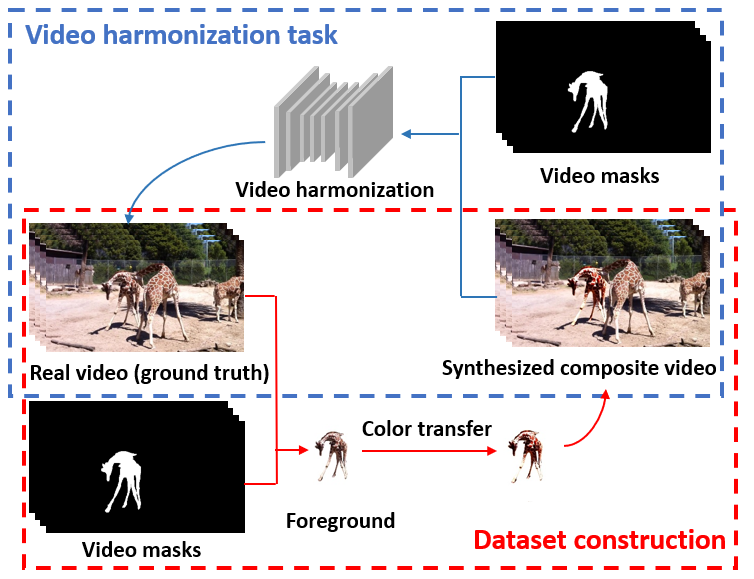}
    \caption{Illustration of video harmonization task (blue arrows) and dataset construction process (red arrows).}
    \label{fig:dataset_construction}
\end{figure}

As a closely related task, image harmonization has attracted growing research interest. Traditional image harmonization methods \cite{2006Color,2007Using,2010Multi,2012RUSHMEIER} attempted to learn the hand-crafted appearance transform between foreground and background, but they neglected the high-level appearance gap between foreground and background. Recently, several deep learning based image harmonization methods \cite{2020DoveNet,2020Improving,guo2021intrinsic,2021Bargainnet,2020Foreground,ling2021region} have been proposed. They changed the foreground style to be harmonious with the background using deep learning techniques. 

Although deep image harmonization methods have achieved remarkable success, directly applying them to video harmonization by harmonizing each frame separately will cause flickering artifacts \cite{2019Temporally}, which largely downgrades the harmonization quality. Thus, it is imperative to design deep video harmonization method by taking temporal consistency into consideration. To the best of our knowledge, the only deep video harmonization method is \cite{2019Temporally}, which proposed an end-to-end network to harmonize the composite frames while considering the temporal consistency between adjacent frames.

Training deep video harmonization network requires abundant pairs of composite videos and their ground-truth harmonized videos, but manually editing composite videos to obtain their harmonized videos is extremely tedious and expensive. 
Therefore, \cite{2019Temporally} adopted an inverse approach, that is, creating synthetic composite videos from real images. Particularly, they applied the traditional color transfer method \cite{946629} to the foreground of the real image to make it incompatible with the background, leading to the synthetic composite image. Then, they applied affine transformation to foreground and background to simulate the motion between adjacent frames, through which synthetic composite video (\emph{resp.}, ground-truth video) are created based on synthetic composite image (\emph{resp.}, real image). Nevertheless, there is a huge gap between the simulated movement and the complex movement in realistic videos. Moreover, the dataset constructed in \cite{2019Temporally} is not publicly available. 
Different from \cite{2019Temporally}, we create synthetic composite videos based on real video without sacrificing realistic motion. Specifically, we apply color transfer based on lookup table (LUT) \cite{jiang2021ssh} to the foregrounds of all frames. We construct our video harmonization dataset named HYouTube based on YouTube-VOS 2018 \cite{2018YouTube}, leading to 3194 pairs of synthetic composite videos and real videos, which will be detailed in Section~\ref{dataset_construction}.

\section{Dataset Construction}
\label{dataset_construction}

In this section, we will describe the process of constructing our dataset HYouTube based on the large-scale video object segmentation dataset YouTube-VOS 2018 \cite{2018YouTube}. Given real videos with object masks, we first select the videos which meet our requirements and then adjust the foregrounds of these videos to produce synthetic composite videos.

\subsection{Real Video Selection}
\label{sec:video_select}
Since constructing video harmonization dataset requires foreground masks and the cost of annotating foreground masks is very high, we build our dataset based on the existing large-scale video object segmentation dataset
YouTube-VOS 2018 \cite{2018YouTube}. YouTube-VOS contains 4453 YouTube video clips and one video clip is annotated with the object masks for one or multiple objects. Each second has 6 frames with mask annotations and we only utilize these annotated frames. 
Then, for each annotated foreground object in each video clip, if there exist more than 20 consecutive frames containing this foreground object, we save the first 20 consecutive frames with the corresponding 20 foreground masks as one video sample. After that, we remove the video samples with foreground ratio (the area of foreground over the area of the whole frame) smaller than 1\% to ensure that the foreground area is in a reasonable range. After the above filtering steps, there are 3194 video samples left.

\subsection{Composite Video Generation}
Based on real video samples, we adjust the appearance of their foregrounds to make them incompatible with backgrounds, producing synthetic composite videos. 
We have tried different color transfer methods \cite{2010Multi,2012RUSHMEIER,2007Using,zhu2015learning} following \cite{2020DoveNet,2017Deep,2019Temporally} and 3D color lookup table (LUT) \cite{Mese2001Look,2011Medical} following \cite{jiang2021ssh} to adjust the foreground appearance. The color transfer methods \cite{2010Multi,2012RUSHMEIER,2007Using,zhu2015learning} need a reference image and adjust the source image appearance based on the reference image appearance, while LUT is \cite{Mese2001Look,2011Medical} a simple array indexing operation to realize color mapping. We observe that applying \cite{2010Multi,2012RUSHMEIER,2007Using,zhu2015learning} requires carefully picking reference images, otherwise the transferred foreground may have obvious artifacts or look unrealistic. Thus, we employ LUT to adjust the foreground appearance for convenience. Since one LUT corresponds to one type of color transfer, we can ensure the diversity of the composite videos by applying different LUTs to video samples. Firstly, we collect more than 400 LUTs from the Internet. Secondly, we calculate their pairwise differences. Specifically, we sample 1000 real video frames. For each real frame, we apply all the collected LUTs to transfer its foreground to obtain composite frames and calculate fMSE between every two composite images as the pairwise difference between two LUTs. We average the pairwise differences over all 1000 real video frames as the final pairwise difference between two LUTs. Finally, we select 100 mutually different LUTs in an iterative approach to enlarge the diversity of synthesized composite videos.
In particular, in each iteration, we find two closest LUTs from the remaining LUTs and remove one of them. This step is repeated until there are 100 LUTs left. Thus, we get 100 candidate LUTs with the largest mutual difference, which are used to transfer the foregrounds of real video samples.

The process of generating composite video samples is illustrated in Figure~\ref{fig:dataset_construction}. Given a video sample, we first select a LUT from 100 candidate LUTs randomly to transfer the foreground of each frame. Lookup table (LUT) records the input color and the corresponding output color, so one LUT corresponds to one color mapping function $f$.
LUT has been applied in a variety of computer vision tasks. An LUT is a 3D lattice in the RGB space and each dimension corresponds to one color channel (\emph{e.g.}, red). LUT consists of $V=(B+1)^3$ entries by uniformly discretizing the RGB color space, where $B$ is the number of bins in each dimension (we set $B=32$ following the convention in image processing field).
Each entry $v$ in the LUT has an indexing color $\mathbf{c}'_v=(r'_v,g'_v,b'_v)$ and its corresponding output color $\tilde{\mathbf{c}}'_v=(\tilde{r}'_v,\tilde{g}'_v,\tilde{b}'_v)$. The color transformation process based on LUT has two steps: \emph{look up} and \emph{trilinear interpolation}. Specifically, given a color value, we first look up its eight nearest entries in the LUT, and then interpolate its transformed value based on eight nearest entries via trilinear interpolation.

The transferred foregrounds and the original backgrounds form the composite frames, and the composite frames form composite video samples. Following \cite{2020DoveNet}, we set some rules to filter out unqualified composite video samples: 1) The transferred foreground should be obviously incompatible with the background; 2) Although the transferred foreground looks incompatible with the background, the transferred foreground itself should look realistic; 3) The albedo of the foreground should remain the same after color transfer. For example, transferring a red car to a blue car is not meaningful for image harmonization~\cite{2020DoveNet}.
Given a real video sample, if the obtained composite video sample after applying one LUT does not satisfy the above criteria, we will randomly choose another LUT again and repeat the transfer process until the obtained composite video sample satisfies the above criteria. 
 
\begin{figure}[t]
\centering    
\subfigure 
{
	\begin{minipage}[t]{0.45\linewidth}
	\centering          
	\includegraphics[scale=0.26]{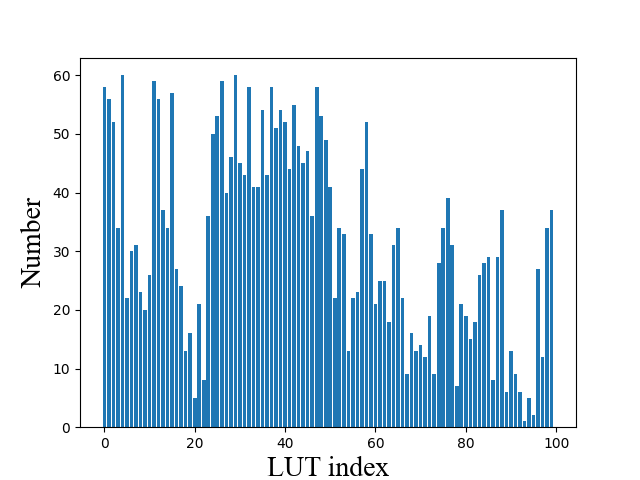}   
	\end{minipage}
}
\subfigure 
{
	\begin{minipage}[t]{0.45\linewidth}
	\centering      
	\includegraphics[scale=0.26]{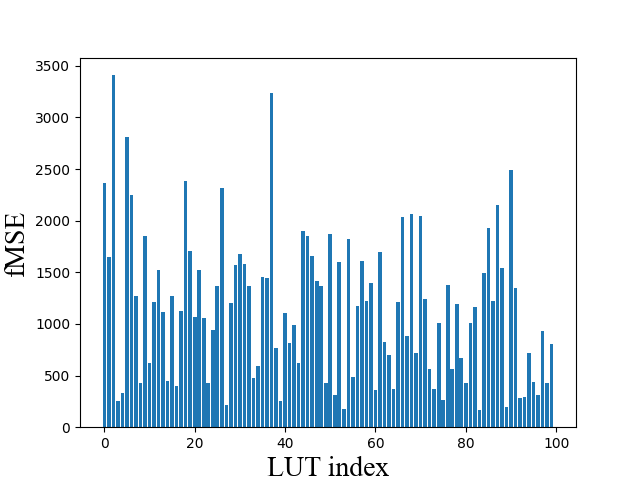}   
	\end{minipage}
}
 
\caption{The left subfigure summarizes the numbers of composite video samples created using different LUTs. The right subfigure summarizes the average fMSE between composite video samples and ground-truth video samples for different LUTs.} 
\label{fig:lut_}  
\end{figure}
\begin{figure*}[t]
    \centering
    \includegraphics[width=0.99\textwidth]{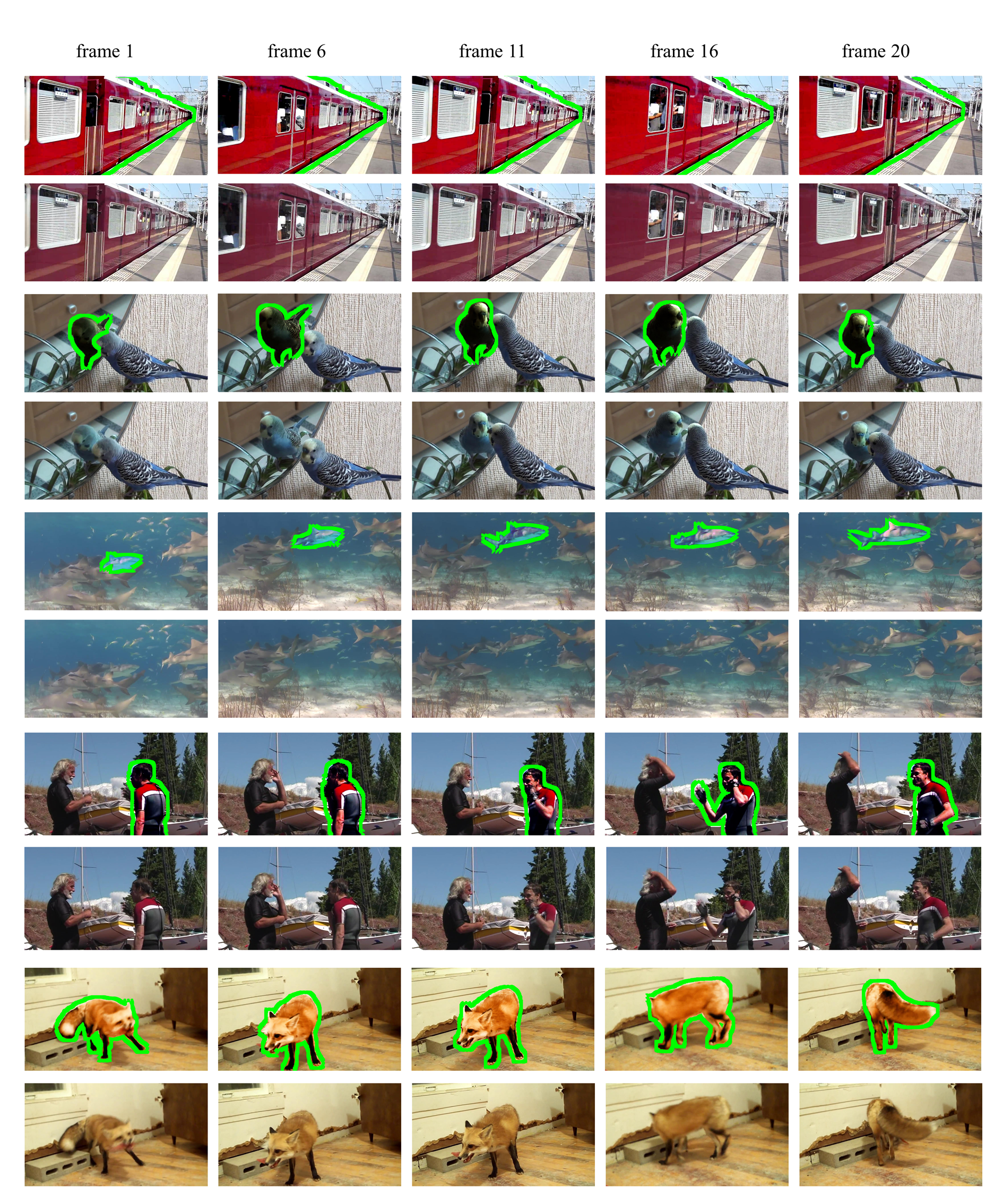}
    \caption{Some example pairs of composite video samples and real video samples. Odd rows are composite video samples and even rows are real video samples. The foregrounds are highlighted with green outlines.}
    \label{fig:dataset_samples}
\end{figure*}

We name our constructed video harmonization dataset as HYouTube. HYouTube dataset includes 3194 pairs of synthetic composite video samples and real video samples. Each video sample contains 20 consecutive frames with the foreground mask for each frame. The numbers of composite video samples created using different LUTs are shown in the left subfigure in Figure \ref{fig:lut_}. We can see that all 100 LUTs have been used, but some LUTs are more frequently used because they are suitable for more real video samples. The average fMSE between composite video samples and ground-truth video samples for different LUTs is shown in the right subfigure in Figure \ref{fig:lut_}. It can be seen that the average fMSE using different LUTs is quite different, which proves the diversity of different LUTs to some extent. Finally, we show some example pairs of composite video samples and real video samples in our HYoutube dataset in Figure \ref{fig:dataset_samples}. 

\section{Real Composite Videos}
\label{sec:real_composite}
We have created pairs of synthetic composite videos and ground-truth real videos in Section \ref{dataset_construction}. 
However, the synthetic composite videos may have a domain gap with real composite videos. To create real composite videos, we first collect 30 video foregrounds with masks from a video matting dataset \cite{sun2021deep} as well as 30 video backgrounds from Vimeo-90k Dataset \cite{2019Video} and Internet. Then, we create composite videos via copy-and-paste and select 100 composite videos which look reasonable \emph{w.r.t.} foreground placement but inharmonious \emph{w.r.t.} color/illumination. We have released these 100 real composite videos for evaluation.

\section{Conclusion}
In this paper, we have constructed a new video harmonization dataset HYouTube which consists of pairs of synthetic composite videos and ground-truth real videos. We have also released 100 real composite videos. The contributed datasets will facilitate the future research in the field of video harmonization. 

{\small
\bibliographystyle{ieee_fullname}
\bibliography{main.bbl}

\begin{thebibliography}{10}\itemsep=-1pt

\bibitem{2011Medical}
F. Bo, F. Zhou, and H. Han.
\newblock Medical image enhancement based on modified lut-mapping derivative
  and multi-scale layer contrast modification.
\newblock {\em IEEE}, 2:696--703, 2011.

\bibitem{2006Color}
Daniel Cohen-Or, Olga Sorkine, Ran Gal, Tommer Leyvand, and Ying-Qing Xu.
\newblock Color harmonization.
\newblock {\em Acm Trans Graph}, 25(3):p. 624--630, 2006.

\bibitem{2021Bargainnet}
Wenyan Cong, Li Niu, Jianfu Zhang, Jing Liang, and Liqing Zhang.
\newblock Bargainnet: Background-guided domain translation for image
  harmonization.
\newblock In {\em ICME}, 2021.

\bibitem{2020DoveNet}
W. Cong, J. Zhang, L. Niu, L. Liu, and L. Zhang.
\newblock Dovenet: Deep image harmonization via domain verification.
\newblock In {\em CVPR}, 2020.

\bibitem{2020Improving}
X. Cun and C.~M. Pun.
\newblock Improving the harmony of the composite image by spatial-separated
  attention module.
\newblock {\em IEEE Transactions on Image Processing}, PP(99):1--1, 2020.

\bibitem{guo2021intrinsic}
Zonghui Guo, Haiyong Zheng, Yufeng Jiang, Zhaorui Gu, and Bing Zheng.
\newblock Intrinsic image harmonization.
\newblock In {\em CVPR}, 2021.

\bibitem{2019Temporally}
H.~Z. Huang, S.~Z. Xu, J.~X. Cai, W. Liu, and S.~M. Hu.
\newblock Temporally coherent video harmonization using adversarial networks.
\newblock {\em IEEE Transactions on Image Processing}, 29:214--224, 2019.

\bibitem{jiang2021ssh}
Yifan Jiang, He Zhang, Jianming Zhang, Yilin Wang, Zhe Lin, Kalyan Sunkavalli,
  Simon Chen, Sohrab Amirghodsi, Sarah Kong, and Zhangyang Wang.
\newblock Ssh: A self-supervised framework for image harmonization.
\newblock {\em arXiv}, 2021.

\bibitem{2007Using}
Jean~Franois Lalonde and A.~A. Efros.
\newblock Using color compatibility for assessing image realism.
\newblock In {\em ICCV}, 2007.

\bibitem{ling2021region}
Jun Ling, Han Xue, Li Song, Rong Xie, and Xiao Gu.
\newblock Region-aware adaptive instance normalization for image harmonization.
\newblock In {\em CVPR}, 2021.

\bibitem{Mese2001Look}
Mese, Murat, Vaidyanathan, and P. P.
\newblock Look-up table (lut) method for inverse halftoning.
\newblock {\em IEEE Transactions on Image Processing}, 10(10):1566--1578, 2001.

\bibitem{946629}
E. Reinhard, M. Adhikhmin, B. Gooch, and P. Shirley.
\newblock Color transfer between images.
\newblock {\em IEEE Computer Graphics and Applications}, 21(5):34--41, 2001.

\bibitem{2020Foreground}
K. Sofiiuk, P. Popenova, and A. Konushin.
\newblock Foreground-aware semantic representations for image harmonization.
\newblock {\em arXiv}, 2020.

\bibitem{sun2021deep}
Yanan Sun, Guanzhi Wang, Qiao Gu, Chi-Keung Tang, and Yu-Wing Tai.
\newblock Deep video matting via spatio-temporal alignment and aggregation.
\newblock In {\em CVPR}, 2021.

\bibitem{2010Multi}
K. Sunkavalli, M.~K. Johnson, W. Matusik, and H. Pfister.
\newblock Multi-scale image harmonization.
\newblock {\em Acm Trans Graph}, 29(4):1--10, 2010.

\bibitem{2017Deep}
Y.~H. Tsai, X. Shen, Z. Lin, K. Sunkavalli, X. Lu, and M.~H. Yang.
\newblock Deep image harmonization.
\newblock In {\em CVPR}, 2017.

\bibitem{2018YouTube}
Ning Xu, Linjie Yang, Yuchen Fan, Dingcheng Yue, Yuchen Liang, Jianchao Yang,
  and Thomas Huang.
\newblock Youtube-vos: A large-scale video object segmentation benchmark.
\newblock {\em arXiv}, 2018.

\bibitem{2012RUSHMEIER}
S. Xue, A. Agarwala, J. Dorsey, and H. Rushmeier.
\newblock Rushmeier h.: Understanding and improving the realism of image
  composites.
\newblock In {\em Acm Trans Graph}, 2012.

\bibitem{2019Video}
T. Xue, B. Chen, J. Wu, D. Wei, and W.~T. Freeman.
\newblock Video enhancement with task-oriented flow.
\newblock {\em International Journal of Computer Vision}, 2019.

\bibitem{zhu2015learning}
Jun-Yan Zhu, Philipp Krahenbuhl, Eli Shechtman, and Alexei~A Efros.
\newblock Learning a discriminative model for the perception of realism in
  composite images.
\newblock In {\em ICCV}, 2015.

\end{thebibliography}
}

\end{document}